\def\year{2021}\relax
\documentclass[letterpaper]{article} 
\usepackage{aaai21}  
\usepackage{times}  
\usepackage{helvet} 
\usepackage{courier}  
\usepackage[hyphens]{url}  
\usepackage{graphicx} 
\usepackage{natbib}  
\usepackage{caption} 
\usepackage{times}
\usepackage{graphicx} 
\usepackage{subfigure}
\usepackage{amsmath}
\usepackage{amsfonts}
\usepackage{color}
\usepackage{url}
\usepackage{wrapfig}
\usepackage{booktabs}
\usepackage{microtype}
\usepackage[switch]{lineno}  %

\usepackage{algorithm}
\usepackage{algorithmic}

\usepackage{tabularx}       

\title{MUFASA: Multimodal Fusion Architecture Search for Electronic Health Records}

\author{
     Zhen Xu\textsuperscript{$\ast$}, David R. So\thanks{Equal contribution}, Andrew M. Dai \\
}

\affiliations{
    Google Research, Mountain View, USA
    \{zhenxu, davidso, adai\}@google.com
}

\begin{document}

\maketitle

\newcommand{\methodname}{MUFASA}
\newcommand{\edit}[1]{{\color{red}<#1>}}

\begin{abstract}
One important challenge of applying deep learning to electronic health records (EHR) is the complexity of their multimodal structure. EHR usually contains a mixture of structured (codes) and unstructured (free-text) data with sparse and irregular longitudinal features -- all of which doctors utilize when making decisions.  
In the deep learning regime, determining how different modality representations should be \textit{fused} together is a difficult problem, 
which is often addressed by handcrafted modeling and intuition.
In this work, we extend state-of-the-art neural architecture search (NAS) methods and propose MUltimodal Fusion Architecture SeArch (\methodname) to simultaneously search across multimodal fusion strategies and modality-specific architectures for the first time. 
We demonstrate empirically that our {\methodname} method outperforms established unimodal NAS on public EHR data with comparable computation costs. In addition, {\methodname} produces architectures that outperform Transformer and Evolved Transformer. Compared with these baselines on CCS diagnosis code prediction, our discovered models improve top-5 recall from 0.88 to 0.91 and demonstrate the ability to generalize to other EHR tasks. Studying our top architecture in depth, we provide empirical evidence that {\methodname}'s improvements are derived from its ability to both customize modeling for each data modality and find effective fusion strategies.
\end{abstract}

\section{Introduction}
\label{introduction}

In recent years, hospitals have begun adopting electronic health record (EHR) systems \cite{AdDe15}. This digitization of large amounts of medical data offers an unprecedented opportunity for deep learning 
to improve healthcare, such as by predicting diagnoses \cite{LiKa15, MiLi16}, reducing healthcare costs \cite{BaSa14, KrAf14}, and modeling the temporal correlation among medical events \cite{ChPu18, XuZh20}. However, EHR data's intrinsic longitudinal and multimodal nature adds distinct complexity that is absent from common academic datasets, such as ImageNet and WMT, that are often used to develop machine learning models.

\begin{figure}[!t]
  \centering
    \includegraphics[width=\columnwidth]{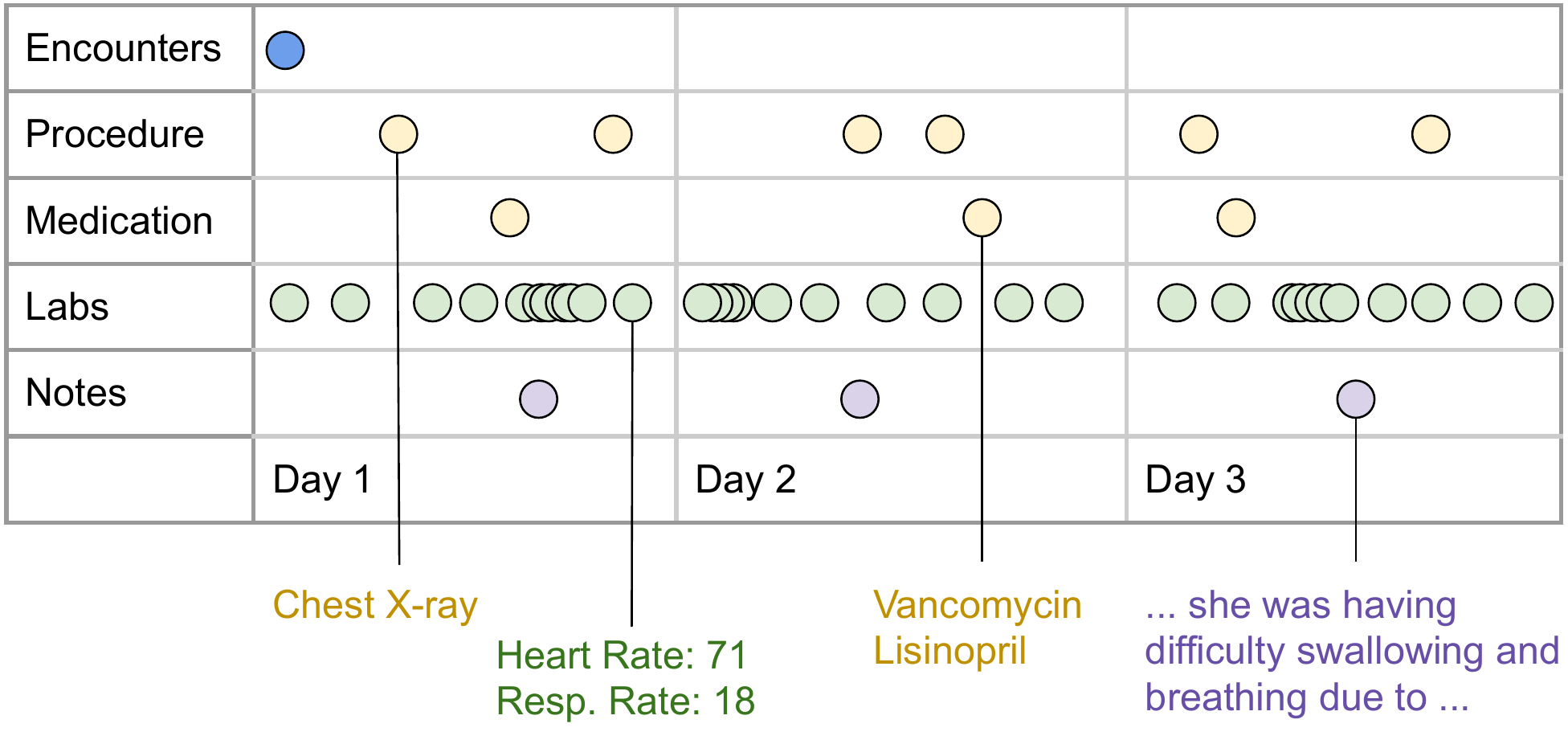}
\caption{An illustrative example of a patient's record. It contains multiple feature modalities, including categorical features, continuous features and clinical notes across time. }
\label{fig:example_ehr}
\end{figure}

Reflecting the complexity of real-world medical information, EHR data contains multiple modalities, both structured (codes and labs) and unstructured (free-text) (Figure~\ref{fig:example_ehr}). For instance, EHR usually contains: (1) contextual features, such as patient age and sex; (2) longitudinal categorical features, such as procedure codes, medication codes, and condition codes; (3) longitudinal continuous features, such as blood pressure, body temperature, and heart rate; and (4) longitudinal free-text clinical notes, which are often lengthy and contain a lot of medical terminology. 
These data types differ not only in feature spaces and dimensionalities, but also in data generation processes and measurement frequencies. For example, lab tests and procedures are ordered at the physician's discretion, while blood pressure and body temperature can be monitored on an hourly basis. 

\begin{figure*}[!htb]
  \includegraphics[width=\textwidth]{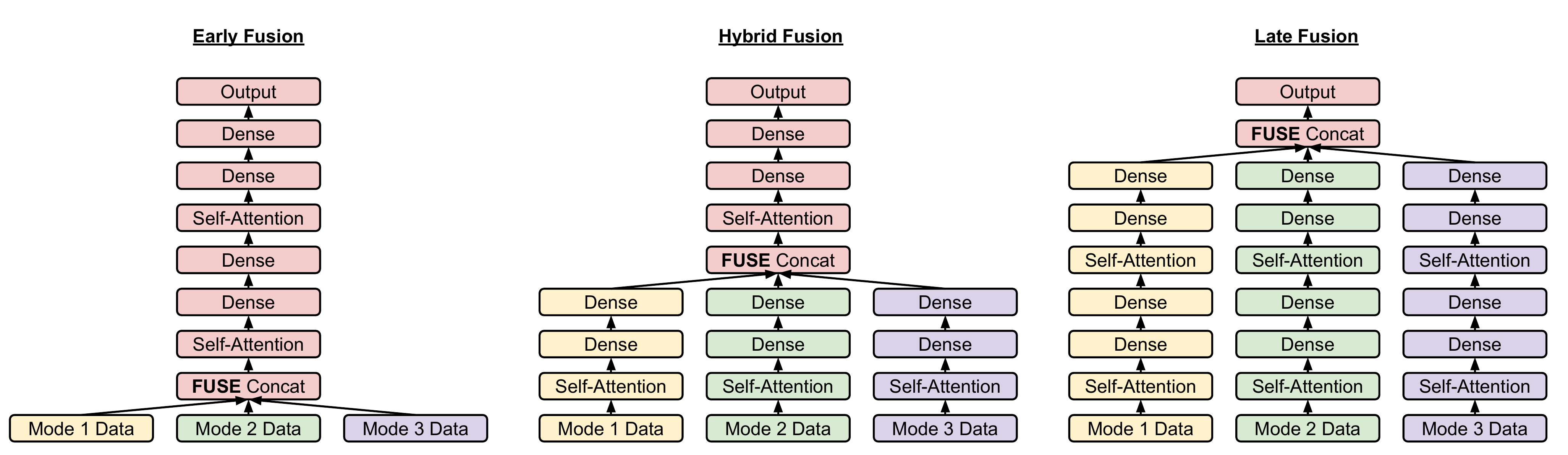}
  \caption{Example of different fusion strategies using a Transformer base architecture. In each depiction, yellow represents modeling only applied to Mode 1, green Mode 2, and purple Mode 3. Red represents modeling applied to all three modes jointly.}
  \label{fig:fusion_example}
\end{figure*}

Synthesizing complimentary information across these multiple modalities allows doctors to make higher quality decisions. For instance, lab tests (continuous features) provide detailed information about a patient's physiological condition, while diagnosis codes (categorical features) capture a system-level view of the patient's state. These modalities have complex interactions; for example, a patient's lab results and diagnoses should be considered when trying to predict what effect a blood pressure medication would have on them. Doctors consider this diverse data, as well as previous doctors' notes, when making decisions.
Thus, modeling these modalities jointly has strong machine learning potential, but needs to be done with care, as adding modalities naively risks making overall model performance worse \cite{RaTa17, BaAh18}. Three questions that guide multimodal modeling are: (1) What model architecture best suits a given modality? For example, convolutional architectures are commonly applied to images, while recurrent neural networks are typically used for temporal data. These decisions are usually based on researchers' intuitions.
(2) Which modalities should be \textit{fused} together? 
``Fusion" refers to the joint modeling of multiple modalities at once by combining their feature embeddings; popular deep learning fusion operations include addition and concatenation.
An example policy is given by \citet{NeWo15}, who argue that highly correlated modalities should be fused together. 
(3) When we do perform modality fusion, at what point in modeling should it occur? Available \textit{fusion strategies} are data-level or early fusion \cite{VaOl16, RaOr18}; intermediate-level or hybrid fusion \cite{LiLi14, SiSi14}; and classifier-level or late fusion \cite{KaBo16, SiZi14} (See Figure \ref{fig:fusion_example} for an example of each strategy).

Thus far, these questions have been addressed by expert hand-designed architectures, which vary from task to task. In this work, we aim to create a generalizable framework to automatically perform this multimodal modeling without meticulous human design. To do this, we look towards \textit{neural architecture search} (NAS) \cite{yao1999evolving}, which has recently produced state-of-the-art results on academic datasets \cite{ReAg19}. However, existing NAS works have primarily focused on unimodal data, and so our focus on applying these techniques to real-world EHR data requires us to offer solutions to tackle the complexities of multimodality. 

We propose \textbf{MU}ltimodal \textbf{F}usion \textbf{A}rchitecture \textbf{S}e\textbf{A}rch (\methodname), which expands the contemporary NAS paradigm to simultaneously optimize multimodal fusion strategies; that is, we jointly search for multiple independent modality-specific architectures, as well as the fusion strategy to combine those architectures at the right representation level. 
We base our searched models on the Transformer \cite{VaSh17} because recent works have shown it can implicitly leverage EHR's internal structure \cite{ChXi18, ChXu19}. Our experimental results show that our discovered {\methodname} models outperform Transformer, Evolved Transformer, RNN variants, and models discovered using traditional NAS, on public EHR data. Specifically, compared with Transformer on Clinical Classifications Software (CCS) diagnosis code prediction, {\methodname} architectures improve test set top-5 recall from 0.8756 to 0.9075. In addition, we empirically demonstrate that {\methodname} outperforms unimodal NAS by customizing each modality specifically -- an ability not available to traditional NAS. Comparing search performance with unimodal NAS on the CCS task, {\methodname} improves validation top-5 recall from 0.9025 to 0.9134 with comparable search costs. What's more, {\methodname} architectures demonstrate more effective transfer to ICD-9, a different EHR task that we do not search on directly. 
Our contributions are summarized as follows:
\begin{itemize}
  \item {\methodname}, the first multimodal NAS that jointly optimizes fusion strategy and modality-specific architectures.
  \item A novel search space that jointly searches unique architectures for distinct modalities and the best strategies to fuse those architectures at the right representation level.
  \item Empirical evidence demonstrating that {\methodname} is superior to traditional NAS for EHR data with comparable computation costs. This includes showing that {\methodname} architectures indeed achieve improvements by customizing modeling for each modality.
\end{itemize}

\section{Related Works}
Recently, machine learning researchers have begun to leverage the multimodal nature of EHR to improve prediction performance \cite{ShHo19}. 
\citet{xuBi18} use both continuous patient monitoring data, such as electrocardiograms, and discrete clinical events to better forecast the length of ICU stays. \citet{QiWu19} propose multimodal attentional neural networks to combine information from medical codes and clinical notes, which improves diagnosis prediction. 
As highlighted by these authors, there are few works integrating streaming and discrete EHR data \cite{xuBi18}, or clinical text and discrete EHR data \cite{QiWu19}. Our work focuses on integrating clinical notes, continuous data, and discrete data all together. Additionally, in contrast to these manually designed multimodal architectures, we explore NAS to automatically learn architectures that leverage the multimodal nature of EHR.

Our work also builds upon \textit{neural architecture search} (NAS). 
Recent results shows that automatically designed deep learning models can achieve state-of-the-art performance on academic benchmarks \cite{ZoLe16, ReAg19}, as well as offer practical usage \cite{tan2019enet}. One-shot NAS methods \cite{bender18one} attempt to radically reduce the amount of compute needed to run searches by not training each candidate individually; here, we use a relatively low compute task and so do not need to employ these methods. Within this NAS field, our work is most closely related to two others. The first is \citet{SoLi19}, who apply NAS to search for a Transformer architecture on NLP data. Our work differs in that we expand our search to include multimodal fusion strategies and modality-specific architectures; in Section~\ref{search_results_section} we compare our search methodology to their unimodal setup and our resulting architectures to the product of their search, the Evolved Transformer. The second comparable work is \citet{PeVi19}, who use architecture search to optimize multimodal feature fusion in image classification models. However, they use off-the-shelf pretrained models as building blocks and only search over their fusion points.
In contrast, we are the first work to jointly optimize multimodal fusion strategies and modality-specific architectures together; this allows us to not only optimize how modalities are fused, but also the type of deep learning computation applied to each modality. Additionally, our focus is on sequence models for EHR data, not convolution-based models for images. 

\section{Methods} \label{sec:methods}
In this section, we briefly describe evolutionary NAS \cite{ReAg19} and the building blocks of our search space. We also describe {\methodname}, our main methodological contribution.

\subsection{Evolutionary Neural Architecture Search}
\label{search_algorithm_section}
We use the \textit{tournament selection} evolutionary architecture search algorithm proposed by \citet{ReAg19}. In this framework, candidate architectures are represented as the \textit{gene encodings} of \textit{individuals}; see Section~\ref{search_space} for a description of these encodings. An initial \textit{population} is created of random or pseudo-random individuals; in our case we use the \textit{warm-start} NAS method \cite{SoLi19} by seeding the initial population with a known strong architecture, the Transformer. From there, evolution begins by assigning every individual in the population a \textit{fitness}. These fitnesses are determined by building the architectures described by each individual's gene encoding and training the resulting models on training data. The models are then evaluated on validation data to determine the individuals' fitnesses. Once fitnesses are assigned, a \textit{tournament} is conducted by sampling $T$ random individuals from the population and selecting the one with the highest fitness to be a \textit{parent}. This parent is \textit{mutated}, with its gene encoding fields randomly changed according to a mutation rate, to produce a \textit{child}. The child is assigned a fitness in the same fashion as the parent. Then another tournament is conducted by sampling $T$ random individuals from the population and having the one with the lowest fitness killed, meaning removed from the population. The newly created and evaluated child is then added to the population in the killed individual's place. This cycle of child creation and weak individual removal is repeated, creating a population of high fitness individuals, which for NAS means strongly performing architectures (Algorithm \ref{evol_algorithm} in Appendix).

\subsection{{\methodname}}
\label{mufasa}

\begin{figure}[!t]
\centering
  \includegraphics[width=0.53\columnwidth]{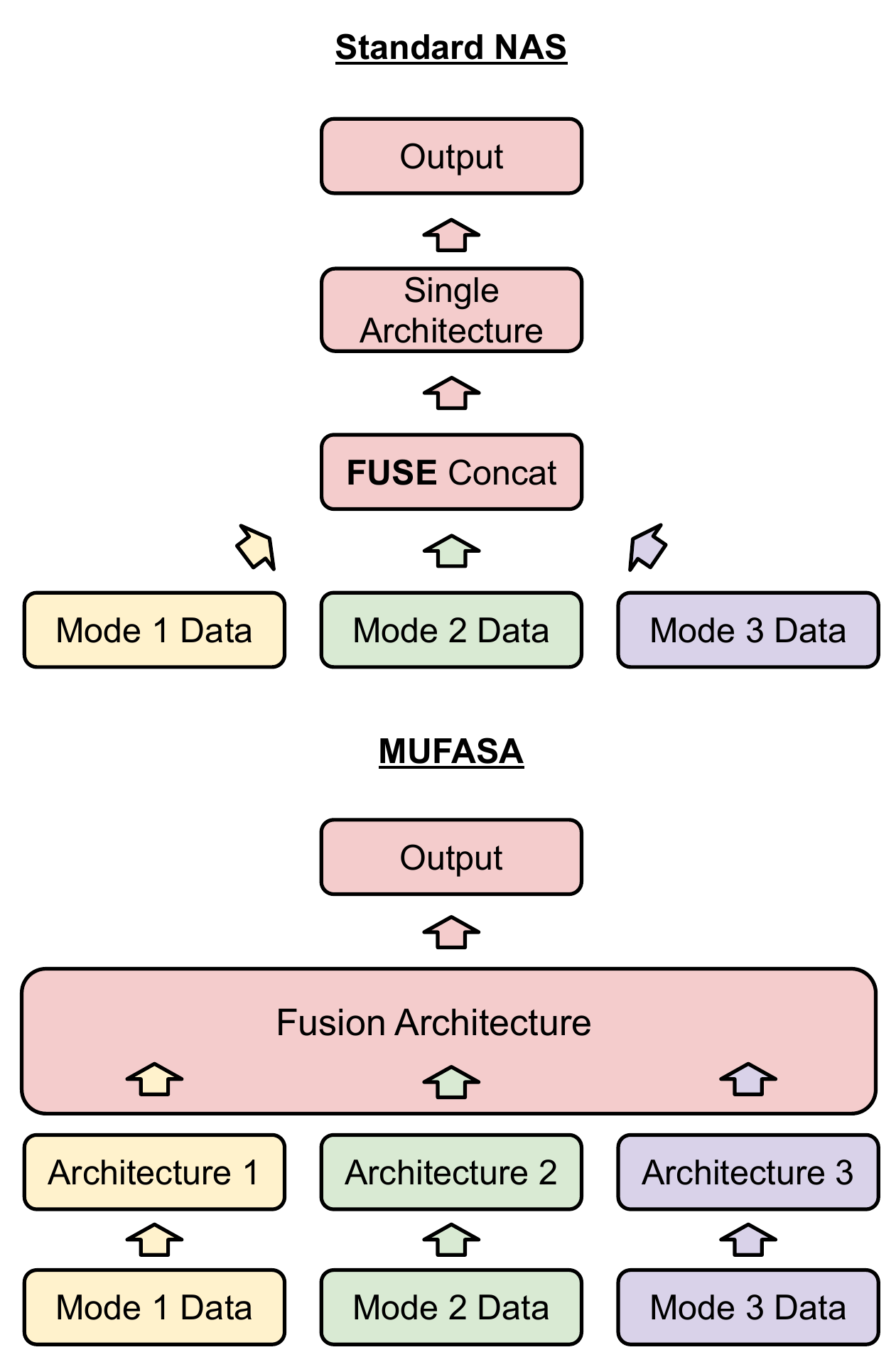}
  \caption{Standard NAS searches for a single unimodal architecture. {\methodname} searches for independent architectures for each data modality, as well as a Fusion Architecture to tie those modal architectures together.}
  \label{fig:uni_vs_multi}
\end{figure}

To adapt to multimodal data, we reformulate the NAS search space to also include fusion strategy search. To do this, instead of searching for a single architecture, we search for several architectures simultaneously: one for each individual data modality and a special \textit{fusion architecture} that is responsible for fusing data modalities together and performing further processing. Put formally, the standard NAS objective is to find an optimal neural network function (architecture) $f_A(x;\theta)$, parameterized by weights $\theta$, that transforms the input $x$ to a representation that is more amenable for a target task. A majority of NAS work, which has focused on unimodal datasets, searches for a single monolithic $f'$. Likewise, several EHR works treat modalities identically, combining all $M$ data modalities together via a simplistic combiner function, such as vector concatenation \cite{LiKa15, RaOr18, ChBa16, li19behrt}, before passing them to one cohesive model (early fusion):
\begin{align*}
f_A(x;\theta) = f'(\operatorname{concat}(x_0, x_1, \dots x_{M-1});\theta)
\end{align*}
Here, we decompose our target architecture into a series of \textit{modality architectures}, $g_i$, that are applied independently to each corresponding $i$th data modality. We additionally define $h$ as the special \textit{fusion architecture} that takes the outputs of each $g_i$ and jointly transforms them into the final output:
\begin{multline*}
 f_A(x;\theta) =
 h(g_0(x_0; \theta_0), \dots g_{M-1}(x_{M-1};\theta_{M-1});\theta_h)
\end{multline*}
During search, {\methodname} searches for the fusion architecture, $h$, and every modality architecture, $g_i$ (Figure~\ref{fig:uni_vs_multi}). This reformulates the search space, distinguishing MUFASA from previous NAS works.
The basis for this reformulation is the notion that for complex data such as EHR, deep learning transformations should be specific to their input modalities; this is represented by the independent modality architectures. As previously mentioned, EHR categorical features, continuous features and clinical notes have different data representations and generative processes; that they would each benefit from distinct types of modeling is intuitive. 
Joint modeling across modalities is also beneficial, but needs to be applied at the right depth; the fusion architecture embodies this mentality. 

In Section~\ref{search_results_section}, we share empirical evidence that supports these ideas that (i) distinct modeling for each modality is beneficial, (ii) proper fusion strategy is critical to model performance, and (iii) the {\methodname} search space is superior to the unimodal NAS space for EHR data, when controlling for search costs.
The utility of {\methodname}'s multi-architecture search is its ability to jointly represent and search over several fusion strategies while performing regular architecture search. In the next subsection, we describe how we construct architectures using typical NAS blocks. Note here that every architecture can be reduced to an identity transformation or, in the case of the fusion architecture, a simple concatenation to perform early fusion. There is a shared parameter budget for the entire model, but there is no explicit constraint on how those parameters can be allocated. 
 
\begin{figure}[t]
  \centering
    \includegraphics[width=0.6\columnwidth]{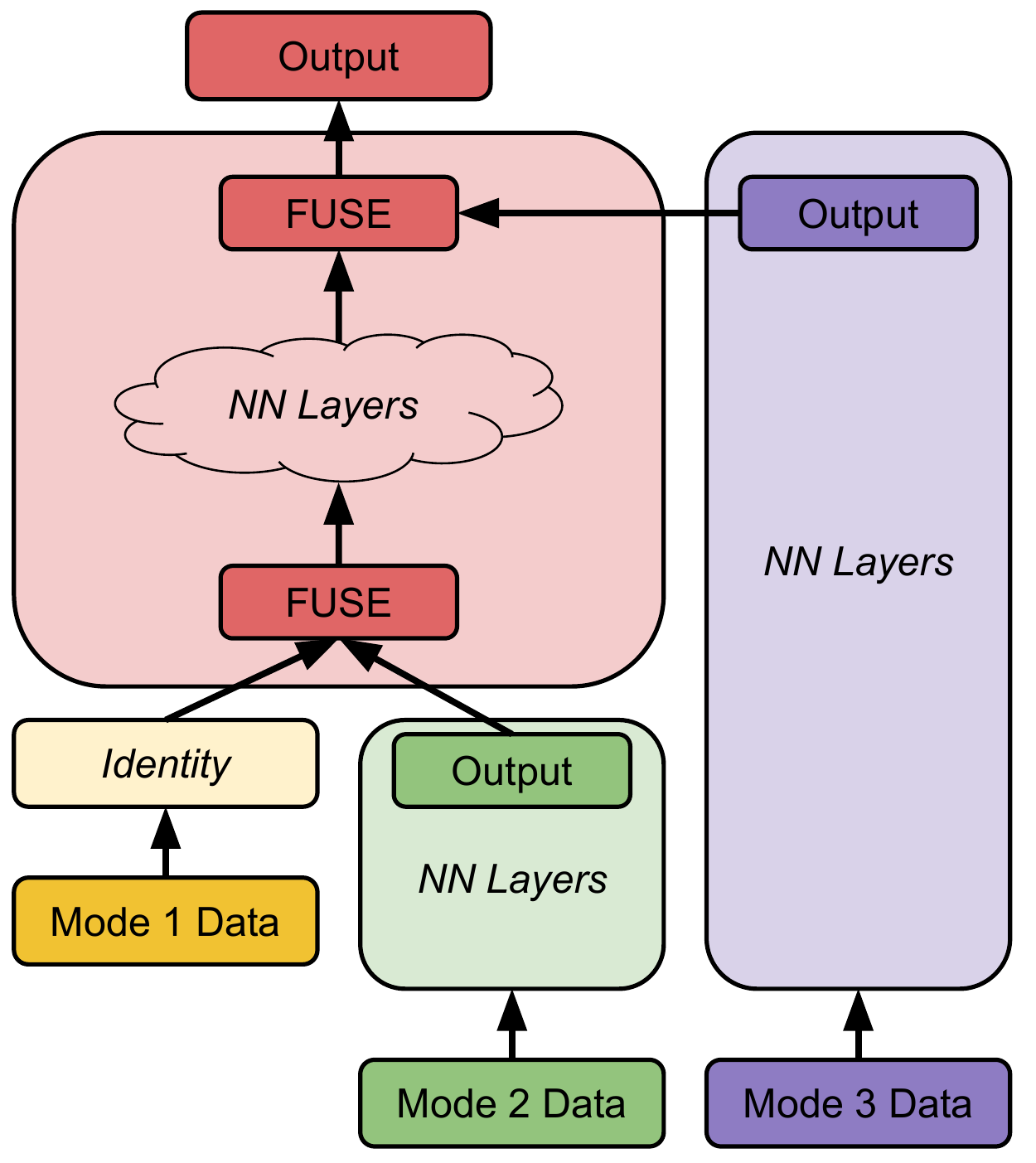}
\caption{{\methodname} example architecture that employs all three fusion types. Mode 1 utilizes early fusion, as its modality architecture is an identity transformation and thus it is fused with Mode 2 before it passes through any neural network (NN) layers. Mode 2 utilizes hybrid fusion, as it first passes through its own non-identity modality architecture before being fused with Mode 1 and transformed by the fusion architecture (in red). Mode 3 utilizes late fusion, as it only receives processing through its independent modality architecture, before being fused with the final output.}
\label{fig:fusion_types}
\end{figure}

\subsection{Architecture Blocks}
\label{search_space}

\begin{figure}[!t]
  \centering
    \includegraphics[width=\columnwidth]{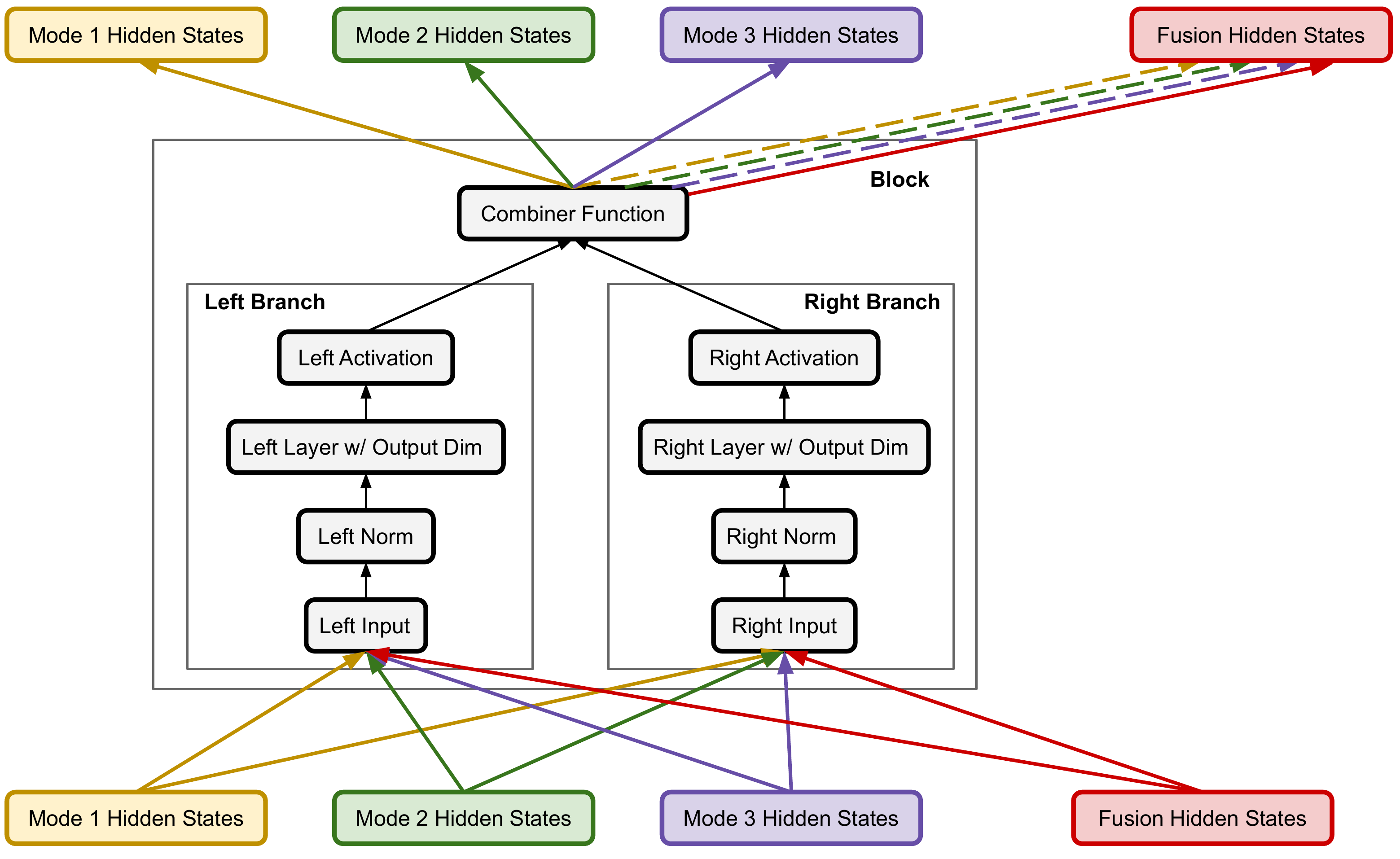}
\caption{Search block structure. In {\methodname}, each modality can access only the previous hidden states from that same modality. The fusion architecture can access fusion architecture states, as well as states from the modality architectures.}
\label{fig:block_structure}
\end{figure}

Similar to previous works, each of the architectures in our search space is  composed of \textit{blocks} (Figure~\ref{fig:block_structure}). Each block receives two hidden state inputs and generates a new hidden state output. The block is a computation unit that transforms each input separately and then combines two transformed outputs together to generate the final block output. The computation applied to each input is called a \textit{branch}. The outputs of both branches are combined via the `combiner function'. The search space for a single block contains $1$ block-level search field (combiner function) and $5$ branch-level search fields (input, normalization, layers, output dimension, and activation) for each of the two branches (10 branch-level fields total). `Input' specifies which previously generated hidden state will be fed into the branch.

Different from previous architecture search work, MUFASA defines two types of blocks, as depicted in Figure~\ref{fig:block_structure}. For modality-specific architecture blocks, only hidden states from the same modality can be inputs. For fusion architecture blocks, both fusion architecture hidden states and modality architecture states can be inputs. Fusion architecture blocks are constructed after the modality-specific blocks have been constructed. These input constraints ensure 1) an independent set of blocks for each modality and 2) that the fusion architecture can access the modality architectures at any representation level, as described by the multi-architecture search space in Section~\ref{mufasa}. Any orphaned hidden outputs are then fused with the model output.
A gene encoding for a single block is represented as \emph{\{left input, left normalization, left layer, left relative output dimension, left activation, right input, right normalization, right layer, right relative output dimension, right activation, combiner function\}}. In total, MUFASA has a search space of $1.76 \times 10^{23}$ models (See Appendix for more vocabulary and search space information).

The fusion architecture can incorporate the modality architectures' outputs at any point in its own architecture; for instance, the fusion architecture can pass the Mode 1 output through its very first neural network layer, and still delay inclusion of the Mode 3 output until the final model layer. It is through being able to freely adjust these two aspects of architecture - parameter allocation and modality inclusion points - that all multimodal fusion strategies can be expressed and searched for via evolution. See Figure~\ref{fig:fusion_types} for an example of how early, hybrid, and late fusion are all achieved. Note, not only can all fusion strategies be represented, but different fusion strategies can be assigned to different modalities. In Section~\ref{search_results_section} we detail the particularly interesting case in which the strongest model we found using {\methodname} applies two different fusion strategies to the same modality (Figure~\ref{fig:best_architectures_fig}).

\section{Experiment Setup}
\subsection{Dataset and Prediction Tasks}
\subsubsection{Dataset}
\label{datasets_section}
We use the Medical Information Mart for Intensive Care (MIMIC-III) \cite{JoPo16} dataset. It contains single-center real-world EHR data for 53,423 hospital admissions, admitted to critical care units from 2001 to 2012 (See Table \ref{table1} in Appendix). We represent patient's medical histories in Fast Healthcare Interoperability Resources (FHIR) format \cite{MaKr16}, as described by \citet{RaOr18}. After data pre-processing, we have 40,511 patients and 51,081 admissions. For all tasks, data is randomly split into train, validation and test sets in an $8:1:1$ ratio.

\subsubsection{Feature Modalities}
\label{feature_modalities_subsection}
We use three feature modalities for the sequence data: (1) Categorical sequence features, including diagnosis and procedure codes; medication request and administration codes; and admission sources. (2) Continuous sequence features, including lab test results and vital signs such as heart rate, respiratory rate, blood pressure, body temperature, and sodium levels, when they are available. (3) Free-text clinical notes. 
Before feeding this data to our models, we embed the categorical sequence features and clinical notes (trained from scratch). We normalize continuous feature values to Z-scores using training set statistics and clamp outliers 10 standard deviations away from the mean. For values that are missing at particular time steps, we use the last observed value for that signal.
The outputs of the searched architectures constitute the sequence representations. After concatenating these representations with additional context features (such as age), we feed the output into dense layers to generate the final task predictions. More details can be found in Appendix Section \ref{appendix_dataset}.

\subsubsection{Prediction Tasks}
Our experiments focus on two diagnosis code prediction tasks at discharge time for each encounter: 
\begin{itemize}
\item \textbf{CCS:}~~~Predicting the primary Clinical Classifications Software (CCS) diagnosis code \cite{ElAn98}. This is a multiclass problem and each hospital encounter has only one primary CCS code. Because there are over 250 possible diagnosis codes, we use top-5 recall (recall@5) as the main evaluation metric.
\item \textbf{ICD-9:}~~~Predicting the International Classification of Diseases, 9th Revision (ICD-9) diagnosis code \cite{SlVe78}. This is a multilabel problem, as one hospital encounter could have several of the 14,000 available ICD-9 diagnosis codes. 
We use AUCPR as the main evaluation metric.
\end{itemize}

\subsection{Baseline Search Algorithms and Model Architectures}
The baseline models that we compare against are the original Transformer \cite{VaSh17}, LSTMs \cite{RaOr18}, attentional bidirectional LSTMs \cite{QiWu19} and the Transformer NAS variant, the Evolved Transformer, which was searched for on translation data. To demonstrate the effectiveness of our {\methodname} search method, we compare it against the same unimodal NAS setup that was used by \citet{SoLi19}, but using our search space vocabulary (Appx. Table~\ref{table:vocab}). This baseline NAS method is not amenable to multimodal inputs and so we concatenate the inputs together before feeding them into each candidate model, as is standard practice in EHR literature \cite{LiKa15, RaOr18, ChBa16, li19behrt}.

\begin{figure*}[!htb]
  \includegraphics[width=\textwidth,height=9cm]{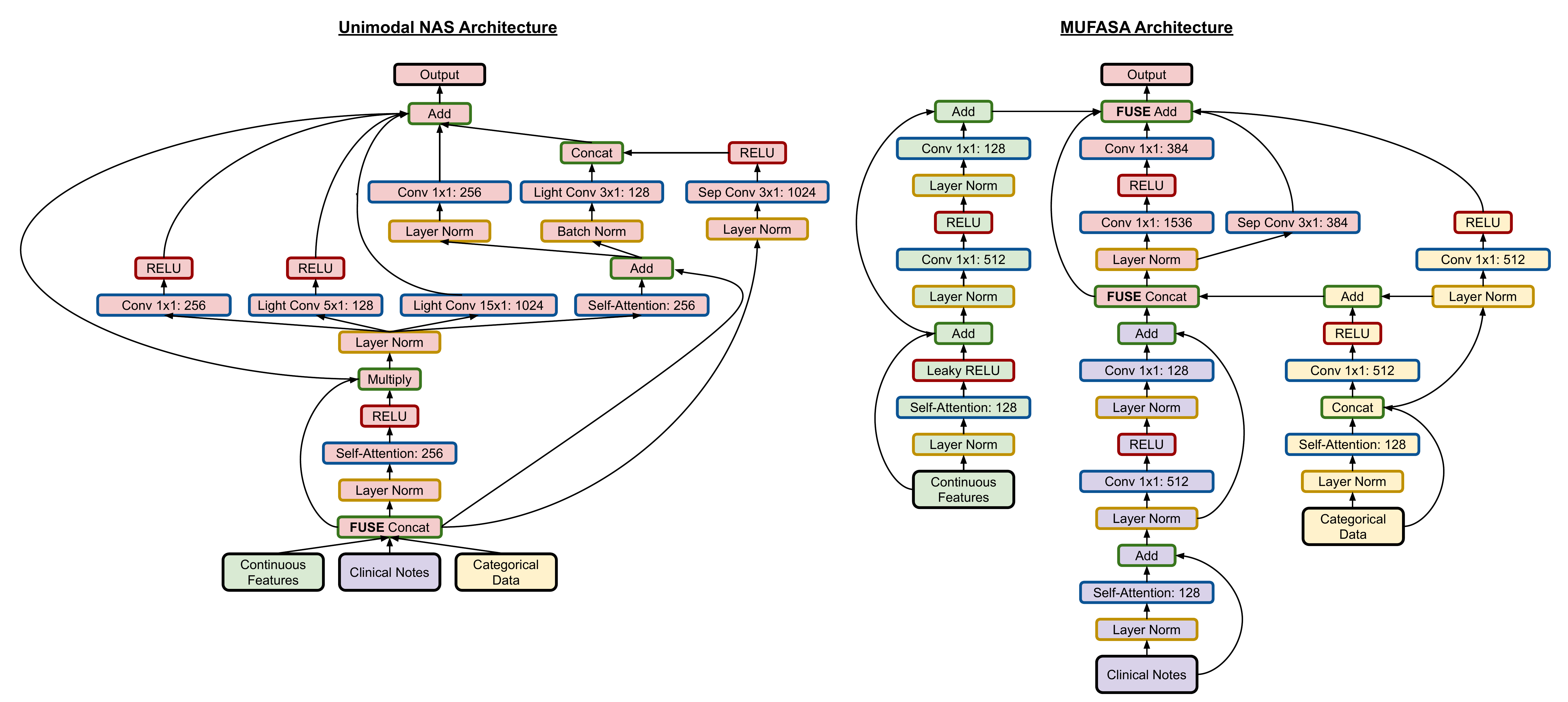}
  \caption{Discovered model architectures from unimodal NAS and {\methodname} on the MIMIC CCS task. The border color of each node represents the type of function applied: yellow for normalization, red for nonlinearity, blue for neural network layer, and green for branch combiner. The color shading for each node represents the data that flows through that node: green is continuous features, purple is clinical notes, yellow is categorical data, and red is multimodal. For {\methodname}, the color shading also indicates which architecture the node is a part of; green, purple and yellow denote the independent modal architectures and red is the fusion architecture. The {\methodname} architecture is much more flexible in terms of what it can represent and applies unique neural architectures to each data modality. In addition, {\methodname} applies different fusion strategies to each modality. For example, the continuous features are processed independently and are only joined with the other modalities at the very end via late fusion. The clinical notes, on the other hand, are processed independently at first, but then are joined with categorical data and processed jointly via hybrid fusion. The most interesting case is the categorical data architecture, which utilizes both hybrid and late fusion.} 
  \label{fig:best_architectures_fig}
\end{figure*}

\subsection{Architecture Search Configuration}
\label{search_config_sub}
We conduct architecture searches on MIMIC CCS. The search configurations are almost identical for {\methodname} and our unimodal search baseline. Both employ the same search space and tournament selection NAS algorithm as described in Section~\ref{sec:methods}. Each search uses 200 CPU workers for evaluating candidate models asynchronously. The population size is 100 and the tournament size is 30. We independently mutate each encoding field with a probability of $1.875\%$ and uniform randomly select its replacement from the possible vocabulary. For each individual architecture search, we train 5000 child models, which in every case appeared to reach convergence. In both searches the parameters for candidate models were not allowed to exceed 76 million. The total search times for both unimodal NAS and {\methodname} are approximately the same: roughly 3 days. 
Unimodal NAS uses the early fusion Transformer to \textit{warm-start} the search. To control for maximum model depth, {\methodname} uses the Transformer hybrid fusion seed (Figure \ref{fig:fusion_example}). See the Appendix for more details.
\footnote{Code is available at https://github.com/Google-Health/records-research/tree/master/multimodal-architecture-search.}

\section{Results}
\label{search_results_section}

\subsection{{\methodname} vs. Unimodal NAS}

We first compare {\methodname} to the equivalent unimodal NAS setup on MIMIC CCS. All relevant configurations, such as compute and hyperparameters, are identical for both searches. We run each search three times and calculate the fitness mean and standard deviation of the best models for each search method.
Compared with unimodal search, {\methodname} improves CCS validation top-5 recall from $0.9025~(0.0041)$ to $0.9134~(0.0034)$; this improvement is statistically significant under independent two-sample t-test with p-value threshold $0.05$.
Figures~\ref{fig:best_architectures_fig} depicts the best architectures from each of the searches. Both architectures are substantially different from the original Transformer seeds, but in very different ways. The unimodal NAS output is ``wider'' than Transformer; it utilizes several wide convolutions and at many points processes the same hidden state through parallel neural network layers. 
Observing this pattern across multiple unimodal NAS searches, we interpret this as the architecture creating multiple ``perspectives" for the same concatenated input, as it is unable to model the modalities individually. 
On the other hand, the {\methodname} architecture scarcely performs parallel computation on the same state, but substantially silos the different modalities and assigns them each unique fusion strategies. For example, the continuous features are processed independently and are only joined with the other modalities at the very end  via  late  fusion.  Clinical  notes,  on  the  other  hand,  are  processed  independently  at  first,  but  then  are  joined  with  categorical  data  and  processed  jointly via hybrid fusion. The most interesting case is the categorical data architecture, which utilizes both hybrid and late fusion.

\begin{table}[ht]
\caption{Test set performance comparison among different model architectures (with different fusion strategies) for the CCS task. We use top-5 recall as the main evaluation metric. We report the mean and standard deviation from 3 runs for each model, with individually tuned hyperparameters.  \textsuperscript{*} denotes that the improvement is statistically significant under independent two-sample t-test with p-value threshold $0.05$.}

\resizebox{\columnwidth}{!}{
\centering
    \label{table:ccs}
    \begin{tabular}{ccc}
    \toprule
     Model                  & Parameters    & Recall@5          \\
    \midrule
     LSTM                   & 38.7M         & 0.8715 (0.0256) \\
     Attentional bidirectional LSTM & 42.7M      & 0.8506 (0.0028) \\
    \midrule 
     Transformer Small (Early)    & 12.7M         & 0.8338 (0.0020)  \\
     Transformer Default (Early)    & 24.5M         & 0.8340 (0.0051)  \\
     Transformer Small (Hybrid)   & 12.7M         & 0.8744 (0.0021) \\
     Transformer Default (Hybrid)   & 24.7M         & 0.8756 (0.0039) \\
     Transformer Small (Late)     & 11.7M         & 0.8727 (0.0041) \\
     Transformer Default (Late)     & 24.8M         & 0.8740 (0.0025) \\
    \midrule 
     Evolved Transformer Small (Early)  & 11.5M   & 0.8372 (0.0032) \\
     Evolved Transformer Default (Early)  & 25M     & 0.8315 (0.0030) \\
     Evolved Transformer Small (Hybrid) & 12.2M   & 0.8722 (0.0042) \\
     Evolved Transformer Default (Hybrid) & 24.6M   & 0.8711 (0.0035) \\
    \midrule 
     Unimodal NAS        & 12.4M         & 0.8841(0.0019) \\
     \textbf{{\methodname}} & \textbf{11.5M}  & \textbf{0.9075}\textsuperscript{*}\textbf{(0.0021)} \\
    \bottomrule
    \end{tabular}
}    
\end{table}

\subsection{Architecture Study}

Having demonstrated that {\methodname} generates better search results than unimodal NAS, we now take a closer look at the architecture it produced and explore what makes that architecture effective. We begin with a comparison to other architectures that have been applied to EHR data, as well as other NAS baselines including the Evolved Transformer and our best unimodal NAS model (Table~\ref{table:ccs}). Note that for the Transformer baselines, which we run at both the default Tensor2Tensor 2 layer size and a size comparable to those found by the searches, fusion strategy has a big impact on performance. Hybrid and late fusion are comparable, but significantly outperform early fusion; fusion strategy being key to model performance is one of the chief motivations for this work. In fact, fusion strategy is so crucial that in this particular case, early fusion performs worse than training on just categorical features or just clinical notes alone (Table~\ref{table:single_modality_importance});
this may be due to the lack of neural network layers that are capable of doing efficient computation across disparate feature representations with different underlying distributions.
The model learned by unimodal NAS outperforms all other baselines that use early fusion; this illustrates the power of EHR-specific modeling. By combining the benefits of both searching for a critically important fusion strategy and modality-specific modeling, {\methodname} produces an architecture that substantially outperforms all other baselines.

\begin{table}[!t]
\caption{\textbf{Testing Independent Modality Modeling:} Input perturbation experiments using {\methodname} architecture.}
\centering
    \label{table:ablation_study}
    \begin{tabular}{cc}
    \toprule
     Perturbation                  & CCS Recall@5        \\
    \midrule
     \textbf{{\methodname} Baseline}        & \textbf{0.9075 (0.0021)} \\
     Forced Early Fusion      & 0.8608 (0.0019) \\
     Shuffle      & 0.8829 (0.0008) \\
     Partial (Categorical $\rightarrow$ Notes)    & 0.9013 (0.0052) \\
     Partial (Notes $\rightarrow$ Categorical)    & 0.8941 (0.0008) \\
    \bottomrule
    \end{tabular}
\end{table}

\begin{table}[!t]
\caption{\textbf{Single Modality Training:} Transformer CCS training on each modality in isolation.}
\label{single_modality_table}
\centering
    \label{table:single_modality_importance}
    \begin{tabular}{cc}
    \toprule
     Modality                  & Recall@5        \\
    \midrule
     Categorical only        & 0.8665 (0.0030) \\
     Continuous only         & 0.7003 (0.0056) \\
     Notes only              & 0.8451 (0.0045) \\
    \bottomrule
    \end{tabular}
\end{table}

The question still stands, however, as to how tailored the {\methodname} architecture is to the individual EHR modalities; its performance on the target task is clearly strong, but how much of that comes from custom modeling for each data modality? To test this, we train three models with the same architecture but with alternative inputs, thereby highlighting the importance of each data modality being fed into its optimized path (Table \ref{table:ablation_study}). 
First, we perform a \textit{forced early fusion}, whereby all input modalities are concatenated and fed into all input paths; this eliminates the isolated modeling of each modality. This experiment shows that the {\methodname} architecture's separate input processing is customized for each individual modality; passing all modalities into every input path together hurts performance.
Second, we randomly \textit{shuffle} the input modalities, passing each modality to a different modality's input path; specifically, categorical features are fed into the continuous features path, continuous features are fed into the clinical notes path, and clinical notes are fed into the categorical features path. This maintains individual modeling for each modality, but changes that modeling from what {\methodname} designates, providing evidence that customized modeling matters. Lastly, we strike a midway point between the first two experiments by performing a \textit{partial} early fusion, adding just one modality to another modality's input path; we randomly try (I) adding categorical features to the clinical notes input path and, in a separate training, (II) adding clinical notes to the categorical features path (for experimental symmetry). Table~\ref{table:ablation_study} shows these improper input routings can cause a statistically significant drop in quality.
The only exception is Partial (Categorical$\rightarrow$Notes); however, this is likely because categorical features are stronger features than clinical notes (Table~\ref{single_modality_table}), so having clinical notes ``share" parameters with categorical features is not as harmful. Note, this still supports the idea that modality-specific processing is sensitive, as the mirrored ``sharing," Partial (Notes$\rightarrow$Categorical), is significantly worse. Jointly, these results confirm our hypothesis that {\methodname}'s improvements come not only from its \emph{independently} modeling modalities, but also its customized modeling \emph{specifically} for those modalities.

To understand if the discovered {\methodname} modeling is effective beyond our target task, we test the generalizability of our discovered architectures on the yet unseen ICD-9 task. As shown in Table~\ref{table:icd9}, {\methodname} demonstrates a statistically significant improvement over the strongest baseline model. The unimodal NAS architecture also seems to generalize, outperforming the strongest early fusion baseline, but is unable to improve over the hybrid fusion Transformer, highlighting its limitations when fusion strategy plays an important role. 

\begin{table}[t]
\caption{\textbf{Generalizability to ICD9 Task:} Comparison between baselines and the CCS-searched models transferred to ICD-9. We train all the same models as in Table \ref{table:ccs}, using the same methodology, but only present the strongest baselines. \textsuperscript{*} denotes the improvement is statistically significant under independent two-sample t-test with p-value threshold $0.05$.
}
\resizebox{\columnwidth}{!}{
\centering
    \label{table:icd9}
    \begin{tabular}{cccc}
    \toprule
     Model                  & Parameters    &    AUCPR          \\
    \midrule
     Transformer Small (Early)    & 89.2M         & 0.3047 (0.0009)  \\
     Transformer Default (Hybrid)   & 101.3M         & 0.3273 (0.0012) \\
     Unimodal NAS        & 89M         & 0.3200 (0.0016) \\
     \textbf{{\methodname}} & \textbf{86M}& \textbf{0.3327}\textsuperscript{*} \textbf{(0.0009)} \\
    \bottomrule
    \end{tabular}
}
\end{table}

\section{Conclusion}
Effective modelling of EHR data has great potential to advance healthcare, from improving diagnoses to suggesting treatments. However, its complex multimodal nature has required human experts to hand-design unorthodox models or use one-size-fits-all models -- an approach that does not scale as medical data becomes richer.
To address this, we proposed {\methodname} to automatically design deep learning architectures that directly account for the uniqueness of different modalities. Our empirical results have shown (1) {\methodname} is superior to unimodal NAS on MIMIC-III CCS; (2) the discovered {\methodname} architectures can outperform commonly used baseline architectures and transfer improvements to other EHR tasks; and (3) the effectiveness of {\methodname} is derived from its ability to specifically model various modalities and find effective fusion strategies. 
Future work can investigate applying {\methodname} to other types of medical data modalities, including medical images, waveforms, and genomics.

\section{Ethics}

We see this work as being a positive step towards the democratization of AI, as it will allow non-experts to develop machine learning models for complex multimodal datasets. Although previous evolutionary NAS works have required hundreds of GPUs/TPUs to reach state-of-the-art performance on the most intensely studied academic datasets, we demonstrate that much less compute can be used ($\sim$ 2 CPU years) to improve performance in applied settings, on very important real-world datasets that do not receive as much attention. Lastly, we hope our contribution of applying machine learning to medical datasets helps advance healthcare for all patients. We only use fully de-identified data from MIMIC-III and follow the data agreement.

\bibliographystyle{IEEEtran}
\bibliography{references}

\newpage

\section{Appendix}
\subsection{More Search Space Information}
 As described in Section \ref{search_space}, our search space is composed of blocks. The encoding for a single block contains $1$ block-level search field (combiner function) and $5$ branch-level search fields (input, normalization, layers, output dimension, and activation) for each of the two branches (10 branch-level fields total). `Combiner function' is the function applied to combine the outputs of both branches. `Input' specifies which previously generated hidden state will be fed into the branch. `Normalization' is the normalization immediately applied to the input, before any other transformation. `Layer' is the neural network layer applied after normalization. `Output dimension' defines the relative output dimension of the layer transformation \cite{SoLi19}.
`Activation' is the non-linearity applied to the output of each layer. Together these components define the entire searchable architecture space. The vocabulary for our gene encodings is shown in Table \ref{table:vocab}.

Because {\methodname} searches for multiple architectures simultaneously, we maintain separate sets of blocks for each individual architecture. Specifically, for our EHR tasks there are 3 distinct modes (Subsection~\ref{feature_modalities_subsection}) and so we maintain 4 distinct block sets: one for each of the 3 independent modality architectures and one for the fusion architecture.  Unlike previous works that leverage similar spaces \cite{ZoVa18, ReAg19}, we do not have the concept of a larger `cell' that is stacked; our dataset is relatively small and so we find having additional stacked cells does not help. 

\subsection{Relationship Between Figures 3-5}
Figures 3-5 all describe the same MUFASA search space, but at varying levels of abstraction. Figure 3 is the highest in abstraction, giving an overview of how the different types of MUFASA architectures (modality and fusion) interact; the modality architectures are independent, and the fusion architecture combines them together. Figure 4 gives a more concrete example with some “layers” exposed, showing how different fusion strategies can be achieved at a macro level. Figure 5 is the most granular, “zooming in” to what a search space block looks like and what inputs different MUFASA blocks can take while preserving the independence assumptions of the modality blocks.

\subsection{Training Details}
All training parameters are optimized on top of the default Tensor2Tensor hyperparameters \cite{VaBe18} for a Transformer trained with an Adam optimizer \cite{KiBa15}. We additionally tune the learning rate schedule and batch size for all model trainings outside of searches. The possible learning rate schedules are constant, linear decay, exponential decay, single cycle cosine decay, and inverse-square-root decay. The possible batch sizes range from 16 to 512. We use a Gaussian process bandit optimization algorithm \cite{DeKr14} to tune hyperparameters to maximize the model performance using evaluation results on the validation set.  
For search, we use the best baseline Transformer hyperparameters for all models; that is a single-cycle cosine decay learning rate schedule without warm-up, batch size 32, and an initial learning rate of $4.23 \times 10^{-4}$. All models are trained until convergence. 

\begin{table}[!ht]
\caption{\textbf{Architecture Vocabulary}}
\centering
    \label{table:vocab}
    \begin{tabular}{lp{5cm}}
    \toprule
     \textbf{Search Field}                      &  \textbf{Vocabulary}         \\ 
    \midrule
     \emph{Layer}                   & \parbox{5cm}{\textbullet \- Standard convolution $s\times1$ for $s\in\{1,3\}$ \\ \textbullet \- Depthwise separable convolution \cite{Sifre14ecolepolytechnique} $s\times1$ for $s\in\{3,5,7,9,11\}$ \\ \textbullet \- Lightweight convolution \cite{WuAn19} $s\times r$ for $s\in\{3,5,7,15\}$ and reduction factor $r \in \{1,4,16\}$ \\ \textbullet \- $n$ head self attention for $n\in\{4,8,16\}$ \\ \textbullet \- Gated linear unit \cite{DaFa17}\\  \textbullet \- Max pooling $s\times1$ for $s\in\{3,5\}$ \\ \textbullet \-  Average pooling $s\times1$ for $s\in\{3,5\}$ \\ \textbullet \- Identity \\ \textbullet \- Dead branch}  \\
    \midrule
     \emph{Activation}           & Relu, Leaky relu \cite{MaHa13}, Swish \cite{RaZo17, elfwing17swish}, None \\
    \midrule
     \emph{Normalization}        & Layer normalization, Batch normalization, None \\
    \midrule
     \emph{Combiner}           & Addition, Concatenation, Multiplication \\
    \bottomrule 
    \end{tabular}
\end{table}

\begin{algorithm}[t!]
\caption{Evolutionary Architecture Search}
\label{evol_algorithm}
\begin{algorithmic}[1]
\REQUIRE Population Size $P$, Tournament Size $T$, Number of Candidates $C$ 
\STATE Population $\leftarrow$ MUTATE(Transformer) $\times P$
\STATE Candidates $\leftarrow $ $\emptyset$
\WHILE {$|$Candidates$| < C $}
\STATE Tournament\_1 $\leftarrow$ SAMPLE(Population, $T$)
\STATE Parent $\leftarrow$ Highest Fitness in Tournament\_1
\STATE Child $\leftarrow$ MUTATE(Parent)
\STATE Child Fitness $\leftarrow$ TRAIN\_AND\_EVAL(Child)
\STATE Tournament\_2 $\leftarrow$ SAMPLE(Population, $T$)
\STATE Dead $\leftarrow$ Lowest Fitness in Tournament\_2
\STATE Replace Dead with Child
\STATE Add Child to Candidates
\ENDWHILE

\RETURN Highest Fitness Architecture in Candidates
\end{algorithmic}
\end{algorithm}

\subsection{Unimodal v.s. {\methodname} Search Configurations}
The two ways in which the unimodal NAS and {\methodname} configurations differ are the number of searchable blocks and starting seeds. Unimodal NAS uses 8 blocks, straightforwardly reimplementing the early fusion Transformer (Figure~\ref{fig:fusion_example}) to \textit{warm-start} the search. {\methodname} uses a Transformer hybrid fusion seed as shown in the middle of Figure \ref{fig:fusion_example}. It has 3 blocks for each of the 3 modality architectures and 5 blocks for the fusion architecture, for a total of 14 blocks. This number of {\methodname} blocks was not tuned, but rather was the minimum number of blocks needed to reimplement a Transformer with hybrid fusion while maintaining the same maximum achievable layer depth as the Unimodal NAS baseline. In preliminary experiments, we found that model depth has a big impact on model performance. So, in the interest of fairness, we enforce all Transformer seeds to have the same depth and a similar number of parameters. Examples of early, hybrid and late fusion seeds are shown in Figure \ref{fig:fusion_example}.

\subsection{Dataset and Data Preprocessing}
\label{appendix_dataset}
We use critical care data from the Medical Information Mart for Intensive Care (MIMIC-III) \cite{JoPo16} in our empirical studies. It contains single-center real-world EHR data for 53,423 hospital admissions, admitted to critical care units from 2001 to 2012. The mean length of stay for all encounters is around 10 days. We represent patient's medical histories in Fast Healthcare Interoperability Resources (FHIR) format \cite{MaKr16}, as described by \citet{RaOr18}. 
Our study cohort is comprised of adult patients hospitalized for at least 24 hours. Detailed statistics are shown in Table \ref{table1}. The extracted data includes encounter information such as admission types, status and sources, diagnosis and procedure codes, observation features such as vital signs and laboratory measurements (See detailed list in Table \ref{table:obs_feature}), medication orders, and free-text clinical notes.

Each patient record is represented as a time series. We normalize continuous feature values to Z-scores using training set statistics and clamp outliers 10 standard deviations away from the mean. Additionally, we embed categorical features. To reduce sequence length and normalize different feature frequencies, we group each feature's values into fixed-length time periods, called \textit{bags}. We aggregate all embeddings or continuous values within the same bag for each feature. In our experiments, we use daily bagging. Lastly, we concatenate the bagged embeddings and continuous values of all features into a single representation for each time period in the patient record. We ignore empty bags, which do not contain any values for the given time period. In this way, the whole patient history is represented as a sequence example. We feed the sequence examples into the searched architectures and their outputs constitute the sequence representations. After concatenating these representations with additional context features such as age, we pass the outputs into dense layers to generate the final task predictions. 

\subsection{Computation Cost for Multiple Modalities}
The number of MUFASA search fields grows linearly with the number of modalities. To justify this increase of search space size, we perform an empirical comparison between MUFASA and unimodal NAS, which has no additional search fields for multiple modalities. MUFASA strongly outperforms unimodal NAS when controlling for the number of searched individuals and both searches cost comparable amounts of time (~3 days).

The compute cost for individual architectures grows linearly with respect to the number of modalities, because additional modalities add additional feature embedding matrices. This is the same for both MUFASA and unimodal NAS. Because we control for the number of candidate model parameters, we did not find a large difference in the training time of different architecture candidates (across both MUFASA and unimodal NAS).

\subsection{Discussion on Architecture Search}
Experimental results in Table~\ref{table:single_modality_importance} shows that the Evolved Transformer, which was searched for using NAS on translation data, does not outperform the baseline Transformer; our interpretation of this is that the inductive biases learned by NAS do not transfer across domains that are significantly disparate, a phenomenon that has been observed in other domains such as computer vision as well \cite{Marco18}. This is alarming because, although academic datasets such as ImageNet and WMT have nice properties for conducting machine learning research, they differ significantly from many real world datasets, such as MIMIC-III. However, while our compute cost for a search is not trivial ($2\sim3$ CPU years in our setup but potentially much faster on GPU or TPU), this is magnitudes smaller than previous NAS works, and yields reasonable results for our particular domain. We believe this is indication that small scale NAS may be well equipped to help model smaller, real world datasets that have not received as much attention from the academic community. 

\subsection{Observation Features}
We list the observation features and their units, which are used in the experiments, in Table \ref{table:obs_feature}.

\begin{table*}[ht]
\label{main_results_table}
\caption{Complete test set performance comparison among different model architectures (with different fusion strategies) for the CCS task. We use top-5 recall as the main evaluation metric. We report the mean and standard deviation from 3 runs for each model, with individually tuned hyperparameters.  \textsuperscript{*} denotes that the improvement is statistically significant under independent two-sample t-test with p-value threshold $0.05$.}

\centering
    \label{table:ccs_big}
    \begin{tabular}{cccccc}
    \toprule
     Model  & Parameters    & Recall@5    & AUCPR  & AUCROC & F1      \\
    \midrule
     LSTM   & 38.7M  & 0.8715 (0.0256) & 0.7114(0.0030) & 0.9868(0.0005) & 0.6383(0.0037) \\
     Attentional bidirectional LSTM & 42.7M & 0.8506 (0.0028) & 0.6985(0.0024) & 0.9862(0.0005) & 0.6253(0.0035)\\
    \midrule 
     Transformer Small (Early)    & 12.7M         & 0.8338 (0.0020) & 0.6874(0.0016) & 0.9868(0.0001) & 0.6218(0.0020) \\
     Transformer Default (Early)    & 24.5M         & 0.8340 (0.0051) & 0.6775(0.0067) & 0.9858(0.0002) & 0.6078(0.0063) \\
     Transformer Small (Hybrid)   & 12.7M         & 0.8744 (0.0021) & 0.7320(0.0050) & 0.9901(0.0005) & 0.6606(0.0045) \\
     Transformer Default (Hybrid)   & 24.7M         & 0.8756 (0.0039) & 0.7382(0.0074)  & 0.9903(0.0004) & 0.6641(0.0065)  \\
     Transformer Small (Late)     & 11.7M         & 0.8727 (0.0041) & 0.7327(0.0038) & 0.9901(0.0004) & 0.6618(0.0050) \\
     Transformer Default (Late)     & 24.8M         & 0.8740 (0.0025) & 0.7302(0.0046) & 0.9898(0.0006) & 0.6575(0.0045)\\
    \midrule 
     Evolved Transformer Small (Early)  & 11.5M   & 0.8372 (0.0032) & 0.6843(0.0021) & 0.9877(0.0002) & 0.6182(0.0021) \\
     Evolved Transformer Default (Early)  & 25M     & 0.8315 (0.0030) & 0.6752(0.0010) & 0.9860(0.0002) & 0.6121(0.0030) \\
     Evolved Transformer Small (Hybrid) & 12.2M   & 0.8722 (0.0042) & 0.7286(0.0044) & 0.9903(0.0004) & 0.6564(0.0030) \\
     Evolved Transformer Default (Hybrid) & 24.6M   & 0.8711 (0.0035) & 0.7289(0.0065)  & 0.9894(0.0005) & 0.6578(0.0042)  \\
    \midrule 
     Unimodal NAS        & 12.4M         & 0.8841(0.0019) & 0.7312(0.0017)  & 0.9899(0.0002) & 0.6570(0.0008) \\
     \textbf{{\methodname}} & \textbf{11.5M}  & \textbf{0.9075}\textsuperscript{*}\textbf{(0.0021)} & \textbf{0.7693}\textsuperscript{*}\textbf{(0.0023)}  & \textbf{0.9932}\textsuperscript{*}\textbf{(0.0003)} & \textbf{0.6913}\textsuperscript{*}\textbf{(0.0037)}  \\
    \bottomrule
    \end{tabular}
\end{table*}

\begin{table}[t]
\caption{\textbf{Full Table for Generalizability to ICD9 Task:} Comparison between baseline Transformers and the CCS-searched models transferred to ICD-9. Here, the same training methodology as Table \ref{table:ccs} is used. All the same models that were trained in that table were also trained for this comparison. \textsuperscript{*} denotes the improvement is statistically significant under independent two-sample t-test with p-value threshold $0.05$.
}
\vspace{3mm}
\resizebox{\columnwidth}{!}{
\centering
    \label{table:icd9_big}
    \begin{tabular}{cccc}
    \toprule
     Model                  & Parameters    &    AUCPR          \\
    \midrule
     LSTM                       & 107.9M         & 0.2957(0.0009)) \\
     Attentional bidirectional LSTM  & 132.3M      & 0.2903(0.0005) \\
    \midrule 
     Transformer Small (Early)    & 89.2M         & 0.3047 (0.0009)  \\
     Transformer Default (Early)    & 101.5M         & 0.3030(0.0008)  \\
     Transformer Small (Hybrid)   & 89.3M         & 0.3257(0.0005) \\
     Transformer Default (Hybrid)   & 101.3M         & 0.3273 (0.0012) \\
     Transformer Small (Late)     & 88.3M         & 0.3227(0.0033) \\
     Transformer Default (Late)     & 101.3M         & 0.3257(0.0005) \\
    \midrule 
     Evolved Transformer Small (Early)  & 88M   & 0.3007(0.0050) \\
     Evolved Transformer Default (Early)  & 101.6M     & 0.3030(0.0008) \\
     Evolved Transformer Small (Hybrid) & 89M   & 0.3177(0.0012) \\
     Evolved Transformer Default (Hybrid) & 101.2M   & 0.3167(0.0012) \\
    \midrule 
     Unimodal NAS        & 89M         & 0.3200 (0.0016) \\
     \textbf{{\methodname}} & \textbf{86M}& \textbf{0.3327}\textsuperscript{*} \textbf{(0.0009)} \\
    \bottomrule
    \end{tabular}
}
\end{table}

\begin{table*}[t]
  \caption{\textbf{Statistics for MIMIC Patient Cohort:} We use the same patient cohort definition and split as \citet{KeRa19}} 
  \begin{tabularx}{\textwidth}{p{7cm} X X} \toprule
  & Train \& validation & Test \\ \midrule
  Number of patients & 40,511 & 4,439 \\
  Number of hospital admissions* & 51,081 & 5,598 \\
  \quad Sex, $n (\%)$ & & \\
  \qquad Female & 22,468 (44.0) & 2,548 (45.5) \\
  \qquad Male & 28,613 (56.0) & 3,050 (54.5) \\
  \quad Age, median (IQR) & 62 (32) & 62 (33) \\
  \quad Hospital discharge service, $n$ $(\%)$ & & \\
  \qquad General medicine & 21,350 (41.8) & 2,354 (42.1) \\
  \qquad Cardiovascular & 10,965 (21.5) & 1,175 (21.0) \\
  \qquad Obstetrics & 7,123 (13.9) & 803 (14.3) \\
  \qquad Cardiopulmonary & 4,459 (8.7) & 519 (9.3) \\
  \qquad Neurology & 4,282 (8.4) & 457 (8.2) \\
  \qquad Cancer & 2,217 (4.3) & 223 (4.0) \\
  \qquad Psychiatric & 28 (0.1) & 4 (0.1) \\
  \qquad Other & 657 (1.3) & 63 (1.1) \\
  \quad Discharge location, $n$ $(\%)$ & & \\
  \qquad Home & 28,991 (56.8) & 3,095 (55.3) \\
  \qquad Skilled nursing facility & 6,878 (13.5) & 794 (14.2) \\
  \qquad Rehab & 5,757 (11.3) & 653 (11.7) \\
  \qquad Other healthcare facility & 3,830 (7.5) & 448 (8.0) \\
  \qquad Expired & 4,420 (8.7) & 462 (8.3) \\
  \qquad Other & 1,205 (2.4) & 146 (2.6) \\
  \quad Previous hospitalizations, $n$ $(\%)$ & & \\
  \qquad None & 40,362 (79.0) & 4,415 (78.9) \\
  \qquad One & 6,427 (12.6) & 721 (12.9) \\
  \qquad Two to five & 3,681 (7.2) & 397 (7.1) \\
  \qquad Six or more & 611 (1.2) & 65 (1.2) \\
  \quad Number of discharge ICD-9, median (IQR)** & 9 (8) & 9 (8) \\ \bottomrule
  \end{tabularx}
  \caption*{* For primary CCS prediction, 1.3\% of these admissions were excluded, where the primary diagnosis corresponded to a non-billable ICD-9 code. \\ ** Includes only billable ICD-9 codes.} 
  \label{table1} 
\end{table*}

\begin{table}[h!]
\caption{\textbf{List of Observation Features and Units}}
\centering
\footnotesize
\begin{tabular}{l l} \\
\toprule 
Observation Name & Units \\
\toprule 
Anion Gap       &       MEQ PER L       \\
Apnea Interval  &       S       \\
Arterial BP Mean        &       MMHG    \\
Arterial BP [Diastolic] &       MMHG    \\
Arterial BP [Systolic]  &       MMHG    \\
Bicarbonate     &       MEQ PER L       \\
Blood flow      &       ML PER MIN      \\
Calcium [Moles/volume] in Serum or Plasma       &        MG PER DL      \\
Calculated Total CO2    &       MEQ PER L       \\
Cardiac output rate & L PER MIN \\
Chloride        &        MEQ PER L      \\
Creatinine      &        MG PER DL      \\
Eosinophils     &       PERCENT \\
Exhaled minute ventilation low  &       L PER MIN       \\
FIO2    &       PERCENT \\
Foley   &       ML      \\
Glucose &       MG PER DL       \\
Heart Rate      &       BMP     \\
Heart Rate      &       BMP     \\
Hematocrit      & PERCENT               \\
Hemoglobin [Mass/volume] in Blood       &        G PER DL       \\
Lymphocytes dif &       PERCENT \\
Magnesium       &       MG PER DL       \\
Monocytes       &       PERCENT \\
NBP Mean        &       MMHG    \\
NBP [Diastolic] &       MMHG    \\
NBP [Systolic]  &       MMHG    \\
Neutrophils urine       &       PERCENT \\
O2 saturation pulseoxymetry     &       PERCENT \\
Oxygen [Partial pressure] in Blood      &       MMHG    \\
PEEP set        &       CM H2O  \\
PH      &       U       \\
Phosphate       &       MG PER DL       \\
Plateau pressure        &       CM H2O  \\
Platelet Count  &       K PER UL        \\
Potassium       &       MEQ PER L       \\
Present Weight (kg)     &       KG      \\
Previous Weight (kg)    &       KG      \\
Respiratory Rate        &       BMP     \\
Sodium  &       MEQ PER L       \\
SpO2    &       PERCENT \\
Tbili   &       MG PER DL       \\
Temperature C (calc)    &       DEG F   \\
Temperature F   &  DEG F                \\
Urea Nitrogen   &        MG PER DL      \\
Urine output    &       ML      \\
Urine output foley      &       ML      \\
Wbc count       &       K PER UL        \\
Weight Change (gms)     &       G       \\
\bottomrule
\end{tabular}
\label{table:obs_feature}
\end{table}

\end{document}